\documentclass{article}
\usepackage{arxiv}
\usepackage[utf8]{inputenc} 
\usepackage[T1]{fontenc}    
\usepackage{hyperref}       
\usepackage{url}            
\usepackage{booktabs}       
\usepackage{amsfonts}       
\usepackage{nicefrac}       
\usepackage{microtype}      
\usepackage{lipsum}		
\usepackage{graphicx}
\usepackage{natbib}
\usepackage{doi}

 \usepackage{framed,multirow}
\usepackage{tabularx}
\usepackage{graphicx}
\usepackage{amsmath}
\usepackage{lscape}
\usepackage{tablefootnote}

\usepackage{amssymb}
\usepackage{latexsym}
\usepackage{cite}

\title{Adaptive feature-fusion neural network for glaucoma segmentation on unseen fundus images}%
\author{ \href{https://orcid.org/0000-0000-0000-0000}{\includegraphics[scale=0.06]{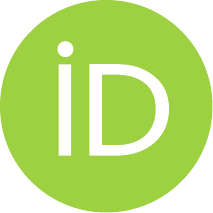}\hspace{1mm}Jiyuan Zhong} \\
	International Medical College,\\
	Chongqing Medical University,\\
	Chongqing, China, 400016\\
	\texttt{zhongjiyuan0516@gmail.com} \\
	\And
	\href{https://orcid.org/0000-0000-0000-0000}{\includegraphics[scale=0.06]{orcid.pdf}\hspace{1mm}Hu Ke}\\
	Department of Ophthalmology,\\
	The First Affiliated Hospital of Chongqing Medical University,\\
	Chongqing, China, 400016\\
	\texttt{42222@qq.com} \\
	\And
	\href{https://orcid.org/0000-0002-5762-395X}{\includegraphics[scale=0.06]{orcid.pdf}\hspace{1mm}Ming~Yan\thanks{Corresponding author:yanmingtop@gmail.com.}  } \\
	Centre for Frontier AI Research (CFAR)\\
        Agency for Science Technology and Research (A*STAR) \\
        1 Fusionopolis Way, 138632, Singapore\\
	\texttt{mingy@cfar.a-star.edu.sg} \\ \\
}
\hypersetup{
pdftitle={A template for the arxiv style},
pdfauthor={Jiyuan~Zhong, Ming~Yan},
pdfkeywords={First keyword, Second keyword, More},
}

\begin{document}
\maketitle

\begin{abstract}
Fundus image segmentation on unseen domains is challenging, especially for the over-parameterized deep models trained on the small medical datasets. To address this challenge, we propose a method named {\em Adaptive Feature-fusion Neural Network (AFNN)} for glaucoma segmentation on unseen domains, which mainly consists of three modules: domain adaptor, feature-fusion network, and self-supervised multi-task learning. Specifically, the domain adaptor helps the pretrained-model fast adapt from other image domains to the medical fundus image domain. Feature-fusion network and self-supervised multi-task learning for the encoder and decoder are introduced to improve the domain generalization ability. In addition, we also design the weighted-dice-loss to improve model performance on complex optic-cup segmentation task. Our proposed method achieves a competitive performance over existing fundus segmentation methods on four public glaucoma datasets.
\end{abstract}





\section{Introduction} 
The most of medical image datasets have quite limited training samples compared to other traditional image datasets. For example, Drishiti-GS~\citep{lowell2004optic}, the fundus image dataset for glaucoma, has only 0.1\textit{K} samples in total. In comparison, the natural image datasets like, MS-Coco~\citep{lin2014microsoft} contains 330\textit{K} training images. Training with limited training data easily leads to inferior performances and tends to be over-fitting in domain generalization tasks. One straightforward way to address the data issue in a low-data regime is to directly feed target domain data into the pretrained model of large natural image datasets and finetune it with the domain-specific feature representation. Unfortunately, the medical images are significantly different from the other natural images in nature. Thus, finetuning strategy is not feasible and the solution is always not optimal if the domain discrepancy between the training and target domain is ignored.

To address the data issue in the fundus image segmentation task, existing transfer learning methods either leverage multiple fundus image sources or conduct data augmentation to increase model domain generalization capability and achieve good performance~\citep{wang2020dofe, liu2021feddg}. In these methods, DeepLab~\citep{chen2017rethinking} or UNet~\citep{ronneberger2015u} are mainly employed as the base models. However, these models with large training parameters are often trained in a fully supervised way, which constrains model learning capability on limited fundus images. Furthermore, the above-mentioned glaucoma segmentation methods fail to consider the domain gap explicitly across their training process as well as the morphology differences between optic-cup and optic-disk. Different from the optic-disk with clearly edge, the optic-cup has a much smaller area and more complex imaging. Simply treating optic-cup and optic-disk equally like existing segmentation methods will lead to a biased optimization. Therefore, designing an adaptive neural network and fully discovering the data's intrinsic property is essential for glaucoma segmentation tasks with limited data. 

To address the aforementioned issues in glaucoma segmentation task, we propose an Adaptive Feature-fusion Neural Network (AFNN), which mainly contains three parts: domain adaptor, feature-fusion network, and self-supervised multi-task learning. Specifically, instead of directly feeding different raw image data from different domains into the glaucoma segmentation network, our AFNN introduces an external domain adaptor to map the diverse data distributions into a common data distribution. This domain adaptation reduces the gaps of divergent domains for deep models on domain generalization task and facilitates more stable training in deep networks. Inspired by the prompt and adaptor~\citep{gao-etal-2021-making} in nature language processing tasks, we introduce the domain adaptor through only tuning a very small number of parameters to help the large pretrained model adaptation on diverse sources meanwhile keeping the prior knowledge of the pretrained model through staged optimization. 

On the top of domain adaptor, we further introduce the feature-fusion network and self-supervised multi-task as the encoder and decoder of AFNN, respectively. On the one hand, the feature-fusion network encourages both multi-layer fusion across different layers and multi-scale fusion in the same network, where the multi-layer fusion helps gradient flow in the deep networks, and multi-scale fusion enriches the representation capability of models. On the other hand, self-supervised multi-task learning seeks to explore more intrinsic correlations between constructed tasks. Concretely, we construct two associate tasks (fundus image reconstruction and domain classification) for the target task of glaucoma segmentation without any additional annotations. Besides the modifications of networks, our AFNN also introduces the weighted-dice-loss and the staged optimization strategy to improve model segmentation performance on optic-disk and optic-cup segmentation tasks, where more weights are assigned to the challenging optic-cup segmentation task.

We conclude the main contributions of our paper as follows,
\begin{itemize}
    \item We develop an adaptive feature-fusion neural network (AFNN) for the glaucoma segmentation task. Through adapting different domains into a common and stable distribution, AFNN largely improves model generalization performance on unseen domains.
    \item We introduce the feature fusion from both multi-levels and multi-scales to enhance model representation capability on generalized segmentation task in the low-data regime.
    \item We develop self-supervised multi-task learning through creating two associate tasks, i.e., fundus image reconstruction and domain classification tasks to improve feature learning for the target task of segmentation. The staged optimization strategy and weighted-dice-loss are demonstrated helpful in further improving model performance on optic-cup and optic-disk segmentation.
\end{itemize}

The rest of our paper is organized as follows. We introduce the related works on domain generalization, and glaucoma segmentation in Section~\ref{Section2:RelatedWorks}, and elaborate details of the proposed network in Section~\ref{Section3:Method}. The experiments and results are reported in Section~\ref{Section4:Experiments}. At last, we gave an ablation study and discussions in Section~\ref{Section5:Discussion}.

\section{Related Works}
\label{Section2:RelatedWorks}
\subsection{Domain Generalization}
Domain generalization applies the machine learning models from pretrained source domains to unseen target domains, which can be mainly grouped into data augmentation and feature alignment methods~\citep{li2017deeper, muandet2013domain}. Specifically, the data augmentation methods aim to enrich the training data with augmentation techniques like the mix-up~\citep{zhang2018mixup} and try to cover the target unseen domain distribution as much as possible~\citep{borlino2021rethinking}. In contrast, the feature alignment domain generalization methods try to learn and align the features rather than raw data through self-learning, contrastive learning, meta-learning, and other regularization strategies~\citep{han2019deep, jing2020self, shorten2019survey}. All these methods improve feature representation capabilities compared to simply combining all the source datasets. Different from these methods, our AFNN employs the domain adaptor to reduce domain divergences, as well as enhances the model representation capability through feature-fusion and self-supervised multi-task learning. 

\subsection{Glaucoma Segmentation}
Glaucoma is a vision-related eye disease and is always diagnosed based on the diameter ratio between the segmented optic cup and optic disk. While, the clinical glaucoma fundus images are mostly captured from different devices, which poses a challenge for the segmentation methods. Some pilot works make efforts to improve glaucoma segmentation on unseen domains. Concretely, Bander~\citep{sym10040087} designs a dense-blob convolution neural network to improve model generalization capability on unseen domains. ASANet~\citep{Affinity2021} improves the affinity alignment and learns invariant patterns across domains by contrastive learning. Different from Bander and ASANet, DoFE~\citep{wang2020dofe} introduces augmented features in its training process, achieving good performance on four glaucoma datasets. Following the same experimental settings of DoFE, ELCFS~\citep{liu2021feddg} further introduces the data privacy protection in the glaucoma segmentation task. Unfortunately, all these glaucoma segmentation works neglect the different importance between the optic-cup and optic-disk segmentation. Our AFNN provides weighted dice loss to solve this imbalance optimization issue in glaucoma fundus images.

\section{Methodology}
\label{Section3:Method}
\begin{figure*}[!t]
\centering
\includegraphics[width=0.8\linewidth]{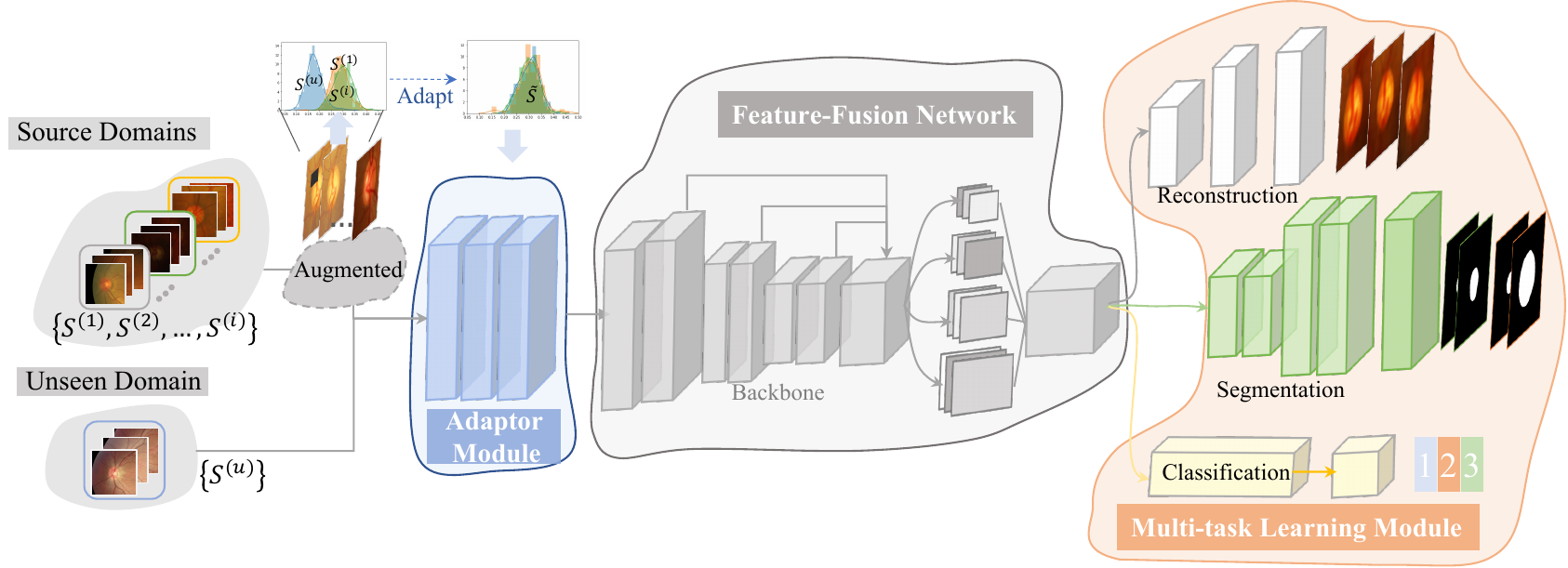}
\caption{Overview of adaptive fusion neural network (AFNN). AFNN mainly contains three modules: the domain adaptor, the feature-fusion network, and the self-supervised multi-task learning module. In particular, the domain adaptor maps a variety of domain distributions into a normalized general distribution. Then with the help of feature-fusion network, AFNN improves its feature learning ability. The last self-supervised multi-task learning further improves model representation ability by learning from the limited training data.}
\label{fig_overview}
\end{figure*}

AFNN aims to learn a generalized representation from the training domains $\{S^{(1)}, S^{(2)}, \cdots, S^{(n)}\}$ with limited data on the unseen domain $S^{(u)}$ for the glaucoma segmentation task. Figure~\ref{fig_overview}  provides an overview of AFNN, which is comprised of three modules: the domain adaptor, the feature-fusion network, and the self-supervised multi-task learning module. We will elaborate on them in the following subsections. 

\subsection{Domain Adaptor}
\label{Setion3.1:Adaptive module}
The medical image datasets typically contain far fewer samples than the regular natural image datasets such as ImageNet~\citep{ILSVRC15}. A straightforward solution to address such data limitation challenge is to train the deep model using every available dataset. Unfortunately, the direct merging of the data collected from various sources is always not an optimal solution resulting in a training conflict since the training data are collected from diverse scanner devices. In addition, the backbone network also faces the domain gap issue, where the source domain used in pretraining differs from the target domain used in finetuning. Those above-mentioned issues seriously prevent the neural networks from learning a generalized representation from the data combinations. 

We introduce the domain adaptor into AFNN to bridge the domain gaps by conducting domain-adaptive learning between different training sources. Inspired by the Adaptor~\citep{houlsby2019parameter} for low-data downstream tasks in natural language processing, our domain adaptor concentrates on reducing domain gaps in low-data scenarios, which adapts different source data to a common data distribution by providing a stable input for the training of neural networks. We also present the two-stage optimization strategy into AFNN to train the neural network in the low-data scenario.
\begin{figure}[!ht]
\centering
\includegraphics[width=\linewidth]{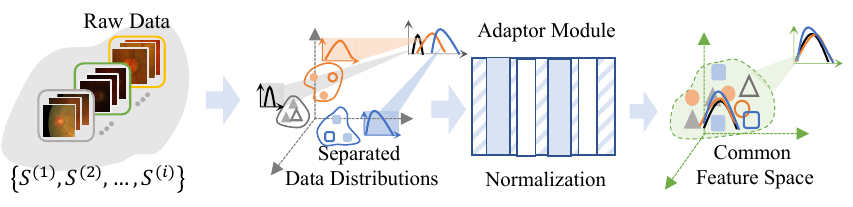}
\caption{Domain-adaptive learning in AFNN. The domain adaptor consists of convolution layers and normalization layers, which maps the raw data to a common distribution and adapts the inputs to the pretrained backbone.}
\label{fig3_adaptive}
\end{figure}

As illustrated in Figure~\ref{fig3_adaptive}, our domain adaptor has two kinds of adaptive blobs. The first instance adaptive blob consists of a convolution layer (Equation~\ref{eq1:ada_conv}) with instance normalization layer (Equation~\ref{eq2:ada_instanceBN}). The second batch adaptive blob consists of a convolution layer and a batch normalization layer (Equation~\ref{eq3:ada_BatchNorm}). Their formulations are listed as follows,
\begin{equation}
\label{eq1:ada_conv}
   z^{(i)} = f_{conv}(S^{(i)} \circ W_{c}),
\end{equation}
\begin{equation}
    \label{eq2:ada_instanceBN}
    \dot{z}^{(i)} =  \frac {z^{(i)} -\dot{\mu}^{(i)}}{ \sqrt { \dot{\sigma^{(i)}}^2+\varepsilon } } ,
\end{equation}
where $W_{c}$ is the weights of the convolutional layer, $\dot{\mu}^{(i)}$ and $\dot{\sigma}^{(i)}$ are the mean and variance of instance feature $z^{(i)}$, correspondingly. As for the batch normalization blob, the convolutional layer also follows Equation~\ref{eq1:ada_conv} but with different normalization, i.e., 
\begin{equation}
\begin{split}
\label{eq3:ada_BatchNorm}
     z^{(i)}      = f_{conv}(\dot{z}^{(i)} \circ W_{c}), \\
    \bar{z}^{(i)} = \frac {z^{(i)} -\bar{\mu}^{(i)}}{ \sqrt{\bar{\sigma^{(i)}}^2+\varepsilon} },
\end{split}
\end{equation}
where $\bar{\mu}^{(i)}$ and $\bar{\sigma}^{(i)}$ are the mean and variance of batch normalization of feature $z^{(i)}$. Concretely, the first instance adaptive blob normalizes the fundus image at the instance-level. While the second batch adaptive blobs expand the normalization to batch-level. As a result, the fundus images from different sources could be transformed and normalized into a common distribution.

Moreover, AFNN introduces a two-stage optimization strategy to alleviate the training issue in low-data scenarios. We observe that the number of the domain adaptor's parameters is far fewer than the number of the backbone's parameters. Therefore, freezing the backbone's parameters ensures the performance of the domain adaptor with limited finetuning data. So, we freeze the backbone parameters and only update the domain adaptor's parameters among different source domains in the first optimization stage. Then, we unfreeze the backbone's parameters and finetune the whole segmentation model in the second optimization stage. We also enlarge the weights of the target segmentation task to enhance model performance on optic-cup and optic-disk segmentation tasks. Consequently, the data from different domains are adapted to a normalized distribution.



\subsection{Feature-fusion Network}
\begin{figure}[!ht]
\centering
\includegraphics[width=\linewidth]{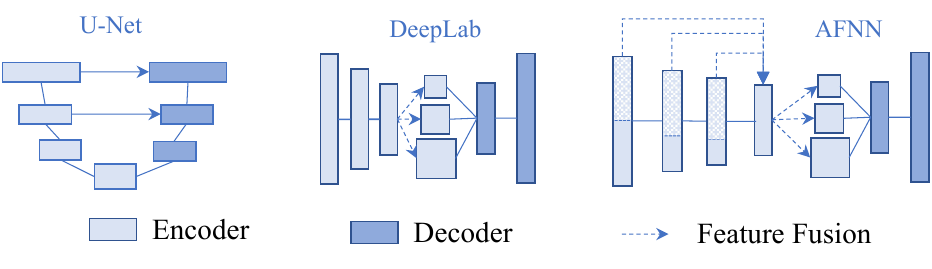}
\caption{The network architectures comparisons of DeepLab, UNet and our feature-fusion network. UNet conducts feature fusion with multiple layers, and DeepLab conducts feature fusion with multiple scales. Our feature-fusion network is a blend of UNet and DeepLab.}
\label{fig4_network}
\end{figure}

We redesign the architecture of the feature-fusion network for AFNN to improve model representation capability in the low-data scenario. Our feature-fusion network integrates DeepLab's good feature representation and UNet's efficient gradient learning ability as a hybrid neural network. It mainly contains two kinds of fusion: multi-level fusion and multi-scale fusion. We compare the architectures of DeepLab, UNet, and our feature-fusion network in Figure~\ref{fig4_network}. 
\label{Setion3.2:FeatureFusion}

Our feature-fusion network has two fusions. The first multi-level feature fusion combines the low-level features with high-level features to enhance the model's representation learning ability in the low-data scenario. Different from U-Net, our AFNN fuses a same dimension of every encoder layers' feature ($f_s^{i} w^i$) as follows,


\begin{equation}
\label{eq4:multi_levFusion}
    z^{l} = 
    \left\{\begin{matrix}
            f_s^{1} w^1 + f_s^{2} w^2 + ... + f_s^{l-1} w^l-1    & i-1 = l, \\ 
            f^{i-1} w^{i-1}                         & i-1 < l,
    \end{matrix}\right.
\end{equation}

The second multi-scale feature fusion conducts representation learning horizontally in the same layer with different scale kernels (\textit{i.e.}, $w_1$, $w_2$), which is formulated as follows,
\begin{equation}
\label{eq5:multi_scalFusion}
    z^{l+1} = ( f^l w_1, f^l w_2, ..., f^l w_k),
\end{equation}

In this way, the feature-fusion network enhances the representation capability effectively by fusing features from vertically multi-level to horizontally multi-scales. 


\subsection{Self-supervised Multi-task Learning}
\label{Setion3.3:Multi-task}
The last self-supervised multi-task learning module of AFNN is proposed for multiple associate tasks (fundus image reconstruction and domain classification) to further improve AFNN's representation learning ability on target glaucoma segmentation task in the low-data scenario. Concretely, AFNN is designed to use the extracted features of its encoder to reconstruct the fundus images for better feature representation. We also leverage the categories of domains as the supervised information to train AFNN for further improvement of AFNN's feature learning. 
We use the same encoder for the fundus image reconstruction and the domain classification in the glaucoma segmentation network. Then, the fused encoder features $z^{l+1}$ are sent to two different decoder branches: the fundus image reconstruction branch (Equation~\ref{eq6:reconstruction}) and the domain classification branch (Equation~\ref{eq7:domainCls}),

\begin{equation}
\label{eq6:reconstruction}
\begin{gathered}
    z_{rec}         = tanh(decoder_{rec}(z^{l+1})),     \\
    \pounds_{rec}   = \left \| x - z_{rec}  \right \|,
\end{gathered}
\end{equation}

\begin{equation}
\label{eq7:domainCls}
\begin{gathered}
    z_{cls}         = softmax(decoder_{cls}(z^{l+1})), \\
    \pounds_{cls}   = - \sum^{n} D \cdot log(z_{rec}), 
\end{gathered}
\end{equation}

We use the $L1$-loss for fundus image reconstruction (Equation~\ref{eq6:reconstruction}) and cross-entropy loss for domain classification (Equation~\ref{eq7:domainCls}). 
All the associated tasks are involved in the training of two optimization stages. We enlarge the segmentation weight in the second stage to ensure that AFNN performs better on the target glaucoma segmentation task.

\begin{figure}[!ht]
\centering
\includegraphics[width=\linewidth]{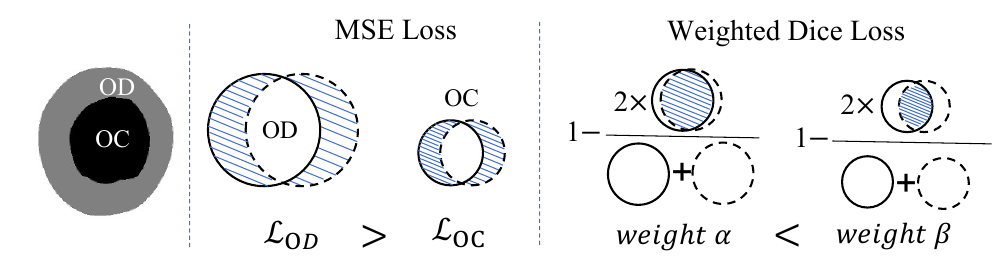}
\caption{Comparison of MSE loss and weighted dice loss.``OD'' and ``OC'' denote the optic-disk and optic-cup correspondingly. ``OD'' and ``OC'' have different intersection areas.}
\label{fig5_lossComparison}
\end{figure}

\textbf{Glaucoma Segmentation Loss} 
Typically, the traditional cost function of the glaucoma segmentation tasks is the dice loss or mean squared error loss (MSE loss). We illustrate the difference between the MSE loss and weighted dice loss in Figure~\ref{fig5_lossComparison}. 

The MSE loss is a metric that measures the difference between ground-truth labels and model predictions. 
However, the optic-disk (OD) has much larger areas than the optic-cup (OC) in fundus images. It leads to the performance dropping in optic-disk segmentation task, where the large-area OD easy gets an over-fitting issue under traditional MSE-based training with the same weights in OC and OD tasks.
To mitigate this sub-optimization issue, we introduce the weighted dice loss (Equation~\ref{eq8:weightDiceLoss}) to place more emphasis on the optic-cup segmentation loss, i.e., 

\begin{equation}
\label{eq8:weightDiceLoss}
\begin{aligned}
    \pounds_{dic} &= \alpha \pounds_{OD} + \beta \pounds_{OC}, \\
            ~     &= \alpha \frac{2\times D_{OD} \cap \hat{Y}_{OD} }{ D_{OD} \cup \hat{Y}_{OD}} + \beta \frac{2\times D_{OC} \cap \hat{Y}_{OC} }{ D_{OC} \cup \hat{Y}_{OC} },
\end{aligned}
\end{equation}
where $\hat{Y}$ is the AFNN's output; $\alpha$ and $\beta$ are the weights of optic-disk and optic-cup, respectively. Our proposed weighted dice loss alleviates the sub-optimization issue by introducing a normalized dice loss with different weights for OC and OD segmentation tasks. Specifically, the losses of large area optic-disk and small area optic-cup are both been normalized by their denominator as shown in Equation~\ref{eq8:weightDiceLoss}. 

\begin{equation}
\label{eq9:Totaloss}
    \pounds = \pounds_{rec} + \pounds_{cls} + \pounds_{dic}.
\end{equation}

Overall, we redesign the segmentation network from domain adaptor, feature-fusion network, and self-supervised multi-task learning as our proposed AFNN, which consists of the multi-task loss and weighted dice loss (Equation~\ref{eq9:Totaloss}. We also offer a staged optimization strategy and a novel weighted-dice-loss to further enhance AFNN's performance in the glaucoma segmentation task.

\section{Experiments}
In this section, we evaluated AFNN on four public glaucoma fundus datasets: Drishiti-GS~\citep{lowell2004optic}, RIME-One-R3~\citep{fumero2011rim}, REFUGE-Train and REFUGE-Test~\citep{orlando2020refuge} to verify the effectiveness of AFNN. All the datasets are collected from different scanner devices with significantly differentiated fundus images, and different datasets are formulated to be individuated domains. 
\label{Section4:Experiments}
\subsection{Datasets and Experimental Settings}
\begin{figure}[!ht]
\centering
\includegraphics[width=\linewidth]{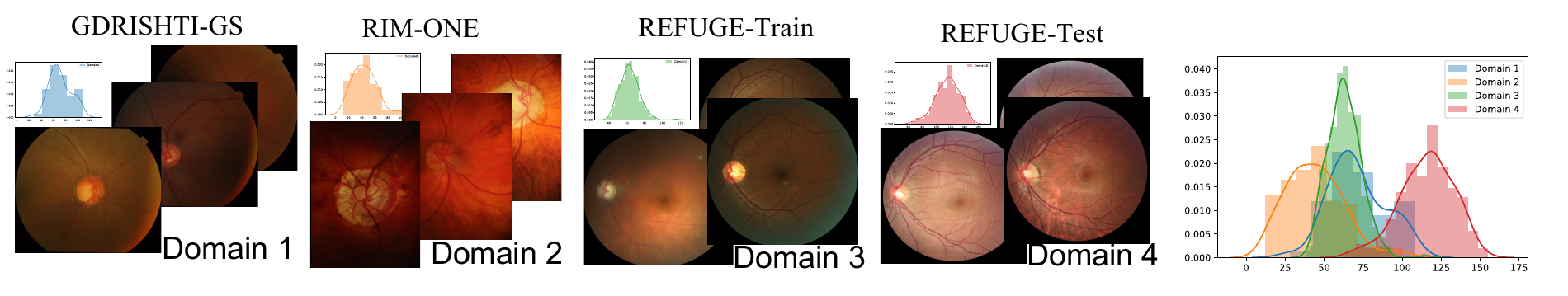}
\caption{Comparisons of different glaucoma datasets. Dataset distribution visualizations are listed on the top-left of each dataset, and the domain gaps of different datasets are shown on the right sub-figure.}
\label{fig6_data}
\end{figure}

From Figure~\ref{fig6_data}, we can first observe the clear differences across different glaucoma datasets (e.g., the lightness, contrast, and resolution). Then, we visualized the distribution of each dataset on its top-left sub-figure. At last, an overview comparison of different datasets is shown on the right of Figure~\ref{fig6_data}. We can find all datasets follow the Gaussian-like distribution except for the Drishiti-GS dataset.

Following the settings of DoFE~\citep{wang2020dofe} and ELCFS~\citep{liu2021feddg}, we also refer each glaucoma dataset as a certain domain (Figure~\ref{fig6_data}). 
Moreover, domain 1 (Drishiti-GS) and domain 2 (RIME-ONE) follow the official training and test splitting. Domain 3 (REFUGE-Train) and domain 4 (REFUGE-Test) strictly follow the splitting of DoFE by the ratio of $1:4$. Note, only the training sets of available domains are fed into AFNN for training, and only the test set of the unseen domain is performed for evaluation. 

\begin{table}[!ht]
\caption{The summary of fundus image datasets.}
\label{tab1:dataset}
\begin{center}
\begin{tabularx}{0.55\linewidth}{rcccc}
\hline
\textbf{Domains} & \textbf{Domain 1} & \textbf{Domain 2} & \textbf{Domain 3} & \textbf{Domain 4} \\ \hline
Training & 50    & 99      & 320         & 320  \\
Test    & 51    & 60      & 80          & 80   \\
FOV     & 30    & 34      & n.a.        & n.a  \\
Devices & NM/FA & AFC-210 & Visucam     & CR-2 \\ \hline
\end{tabularx}
\end{center}
\end{table}

We summarized the experimental dataset information in Table~\ref{tab1:dataset}. All glaucoma datasets are the low-data datasets, where the total training data (less than $1K$ samples) are far less than the natural image datasets (\textit{e.g.}, MS-COCO with more than $200K$ labeled images). Moreover, the training samples are significantly distinct from each other, as they are collected from different scanner devices with the varying field of view (FOV).

\subsection{Evaluation Metrics}
To evaluate our model segmentation performance on the unseen domains, we use the dice similarity coefficient (DSC), Hausdorff distance (HD), and average surface distance (ASD) as our evaluation metrics. The computation formula of the first metric DSC is defined as follows,

\begin{equation}
\label{eq10:DSC_metric}
    \mathbf{DSC}(D, Y) = \frac{2 \times D \cap \hat{Y} }{ D \cup \hat{Y} },
\end{equation}
where $D$ and $\hat{Y}$ are the ground-truth label and the prediction, respectively. From Equation~\ref{eq10:DSC_metric}, we can learn the DSC focuses on the internal overlap areas of ground-truth label and prediction. 

In contrast, HD and ASD focus more on the outer edges of the ground-truth label and model prediction. The difference between those two edge criteria is that the HD is a metric that measures the maximum edge distance between model prediction and ground-truth label, and the ASD is a metric that measures the average edge distance. The definitions of HD and ASD are presented as follows, 

\begin{equation}
\label{eq10:HD_metric}
    \mathbf{HD}(D, \hat{Y}) = \max \{h(D,\hat{Y}), h(\hat{Y},D)\},
\end{equation}
where $h(A,B) = \max\limits_{a \in \mathit{S}(A)}\{\min\limits_{b \in \mathit{S}(B)} ||a-b|| \}$, and ``$\mathit{S}(\star) $'' denotes the set of surface pixels of $\star$. Therefore, HD searches for the maximum value of the shortest distances between two edge pixel sets. While, the ASD is a metric that searches for the average value between the two edge pixel sets:

\begin{equation}
\label{eq10:ASD_metric}
    \mathbf{ASD}(D, \hat{Y}) = \frac{1}{|\mathit{S}(D)|+|\mathit{S}(\hat{Y})|} ( h(D,\hat{Y}) +h(\hat{Y},D) ).
\end{equation}
where $ h(A,B) = \sum\limits_{a \in \mathit{S}(A)} \min\limits_{b \in \mathit{S}(B)} ||a-b||$ and $|\mathit{S}(D)|$ denotes the pixel number of set $\mathit{S}(D)$. 

\subsection{Implementation Details}
AFNN is trained with basic data augmentation, including random flip, random noise, lightness adjustment, and region erasing. The training settings follow the work of DoFE~\citep{wang2020dofe} that the original fundus images are first cropped to $800\times800$ ROIs and then been resized down to $256\times256$ inputs. Regarding the training process being split into two stages, the first training stage sets the same coefficients for multiple losses. While the second stage emphasizes the coefficient of the target task's loss. We initialize the learning rate to $4e-5$ with a cosine decay. All experiments are trained on the NVIDIA TITAN GPU with a batch size of 16. 

\section{Results}

\subsection{Dice Similarity Coefficient}
We first employ the segmentation metric DSC  to evaluate AFNN's generalization performance on unseen domains. Table~\ref{tab2:results_dsc} summarizes the comparison results of our AFNN with recent domain generalization segmentation methods. We categorize these segmentation methods into the general domain generalization methods and the glaucoma domain generalization methods. All the evaluations are performed in the same setting selecting one dataset from the four public glaucoma datasets as the unseen domains for evaluation and the rest three train sets as the source domains for training. The glaucoma segmentation task contains two sub-tasks (optic-cup segmentation and optic-disk segmentation), which are reported separately in Table~\ref{tab2:results_dsc}. 

\begin{table*}[!t]
\caption{Comparison of AFNN with recent glaucoma segmentation methods on unseen domains.}
\label{tab2:results_dsc}
\begin{center}
\resizebox{\textwidth}{!}{ %
\begin{tabular}{rcccccccccc}
\hline
\textbf{Tasks} & \multicolumn{5}{c}{\textbf{Optic-Disk Segmentation}} & \multicolumn{5}{c}{\textbf{Optic-Cup Segmentation}} \\ \hline
Unseen Domains & \textit{1} & \textit{2}  & \textit{3} & \textit{4} & \multicolumn{1}{c|}{\textit{Avg.} $\uparrow$ }  & \textit{1} & \textit{2} & \textit{3} & \textit{4} & \textit{Avg.} $\uparrow$  \\ \hline
Mixup~\citep{zhang2018mixup}   & 0.9297 & 0.8678 & 0.9042 & 0.9076 & \multicolumn{1}{c|}{0.9023} & 0.7332 & 0.7122 & 0.8216 & 0.8623 & 0.7823 \\
M-mixup~\citep{verma2019manifold} & 0.9448 & 0.8938 & 0.9217 & 0.9082 & \multicolumn{1}{c|}{0.9171 } & 0.7927 & 0.7541 & 0.8301 & 0.8673 & 0.8111 \\
CutMix~\citep{yun2019cutmix}  & 0.9383 & 0.9197 & 0.9013 & 0.8879 & \multicolumn{1}{c|}{0.9118} & 0.7697 & 0.8102 & 0.8342 & 0.8683  & 0.8206 \\ 
JiGen~\citep{Carlucci_2019_CVPR} & 0.9392 & 0.8591 & 0.9263 & 0.9404 & \multicolumn{1}{c|}{0.9162} & 0.8226 & 0.7068 & 0.8332 & 0.8570 & 0.8049 \\
DST~\citep{zhang2019unseen} & 0.9220 & 0.9077 & 0.9402 & 0.9066 & \multicolumn{1}{c|}{0.9191} & 0.7563 & 0.8080 & 0.8432 & 0.8624 & 0.8175 \\ \hline
ELCFS~\citep{liu2021feddg}  & 0.9537 & 0.8752 & 0.9337 & 0.9450 & \multicolumn{1}{c|}{0.9269} & 0.8413 & 0.7188 & 0.8394 & 0.8551 & 0.8137 \\
DoFE~\citep{wang2020dofe} & 0.9559 & 0.8937 & 0.9198 & 0.9332 & \multicolumn{1}{c|}{0.9256} & 0.8359 & 0.8000 & 0.8666 & 0.8704 & 0.8432 \\ \hline
AFNN~(ours) & \textbf{0.9602} & \textbf{0.9041} & \textbf{0.9437} & \textbf{0.9492} & \multicolumn{1}{c|}{\textbf{0.9393}} & \textbf{0.8549} & \textbf{0.8229} & \textbf{0.8742} & \textbf{0.8726} & \textbf{0.8562} \\ \hline
\multicolumn{11}{l}{\small *The DSC scores are the quantity evaluation results on optic-disk and optic-cup segmentation tasks.} \\
\end{tabular}}
\end{center}
\end{table*}

Mixup~\citep{zhang2018mixup}, M-mixup~\citep{verma2019manifold} and CutMix~\citep{yun2019cutmix} are three of the general domain generalization methods. They improve model generalization capability by learning from the augmented data of mixed domains. Concretely, with the help of high-level feature mixup instead of raw image mixup, M-mixup and CutMix achieve $2\%$ and $3\%$ higher performance than the vanilla Mixup in average DSC scores. Moreover, JiGen~\citep{Carlucci_2019_CVPR} improved its model representation learning with self-supervised signals and supervised concept information, which also gets better performance than the vanilla Mixup on both sub-tasks. DST~\citep{zhang2019unseen} further improves the model generalization capability by introducing more data augmentation strategies and achieves a $0.9191$ DSC score on the optic-disk segmentation task. At the same time, its performance is still slightly inferior to CutMix on the optic-cup segmentation task.

Different from the aforementioned methods, ELCFS~\citep{liu2021feddg} and DoFE~\citep{wang2020dofe} optimize their segmentation frameworks for the glaucoma dataset. In detail, ELCFS improves the vanilla Mixup segmentation framework by interpolating on the Fourier transformed features instead of raw images. As demonstrated in Table~\ref{tab2:results_dsc}, we can also observe that ELCFS achieves a $0.9269$ DSC score and outperforms the mixup-based methods on the optic-disk segmentation task. However, ELCFS still is a Mixup-based approach with feature interpolating, which leads to an edge blur and performance decline on the challenging optic-cup segmentation task. Instead of interpolating-based augmentation strategies, DoFE improves model generalization by introducing feature-fusion from a domain knowledge pool, which greatly improves the optic-cup segmentation performance to $0.8432$. Based on the idea of feature-fusion, our AFNN further introduces the domain adaptor and self-supervised multiple-task learning module into our segmentation framework to solve the edge blur and low-data training issues, which achieves the best performances on both optic-cup and optic-disk segmentation tasks with around $1\%$ improvement over the existing best glaucoma segment methods ELCFS and DoFE.

\subsection{Hausdorff Distance and Average Surface Distance}
Besides the area-related metric DSC, we also employ the edge-related metrics Hausdorff distance (HD) and average surface distance (ASD) for evaluation that focuses on the performance of segmentation edges instead of overlapped areas.

\begin{table*}[!h]
\begin{center}
\caption{The Hausdorff distance (HD) evaluation results.}
\label{tab:HD}
\resizebox{\textwidth}{!}{
\begin{tabular}{rccccccccccc}
\hline
\textbf{Methods} & \textbf{Metrics} & \multicolumn{5}{c}{\textbf{Optic-Disk Segmentation}}  & \multicolumn{5}{c}{\textbf{Optic-Cup Segmentation}}  \\ \hline
Unseen Domains & \multicolumn{1}{c}{-} & \textit{1} & \textit{2}  & \textit{3} & \textit{4}  & \multicolumn{1}{c|}{\textit{Avg.} $\downarrow$} & \textit{1} & \textit{2}  & \textit{3} & \textit{4} & \textit{Avg.} $\downarrow$ \\ \hline
JiGen~\citep{Carlucci_2019_CVPR} & \textit{HD} & 16.54 & 24.14 & \textbf{11.35} & 12.57 & \multicolumn{1}{c|}{23.01} & 35.42 & 25.63 & 23.74 & 18.59 & 25.85 \\
DST~\citep{zhang2019unseen} & \textit{HD} & 21.84 & \textbf{22.83} & 17.43 & 17.95 & \multicolumn{1}{c|}{20.01} & 43.89 & 24.85 & 21.73 & 14.69 & 26.29 \\
DoFE~\citep{wang2020dofe} & \textit{HD} & 16.21 & 30.23 & 21.45 & 16.10 & \multicolumn{1}{c|}{21.00} & 33.56 & 25.86 & 19.44 & 15.76 & 23.66 \\ \hline
AFNN~(ours)  & \textit{HD} & \textbf{15.62} & 24.16 & 17.46 & \textbf{12.49} & \multicolumn{1}{c|}{\textbf{17.43}} & \textbf{28.87} & \textbf{22.50} & \textbf{18.89} & \textbf{14.43} & \textbf{21.17} \\ \hline
\multicolumn{11}{l}{\small *All HD scores are the average quantity evaluations, and the lower score denotes better performance.} \\
\end{tabular} }
\end{center}
\end{table*}

Table~\ref{tab:HD} summarizes HD evaluation results on four unseen domains. In the optic-disk segmentation task, JiGen and DST are slightly superior to the performance of AFNN on domain 2 and domain 3, separately. This is because our AFNN focuses on enhancing the performance of challenging optic-cup segmentation with a weighted dice loss (Equation~\ref{eq8:weightDiceLoss}). Among all four datasets, our AFNN achieves the best average performance than the compared methods in a more generalized performance. In the challenging segmentation task of optic-cup, AFNN is superior to all compared methods on four evaluation datasets. This results consequently corresponding to our motivation of weighted-dice-loss. Overall, our AFNN constantly achieves superior performance on the average performance of optic-disk and optic-cup segmentation tasks. 

\begin{table*}[!h]
\begin{center}
\caption{The average surface distance (ASD) evaluation results.}
\label{tab:ASD}
\resizebox{\textwidth}{!}{ %
\begin{tabular}{rccccccccccc}
\hline
\textbf{Methods} & \textbf{Metrics} & \multicolumn{5}{c}{\textbf{Optic-Disk Segmentation}} &   \multicolumn{5}{c}{\textbf{Optic-Cup Segmentation}}   \\ \hline
Unseen Domains & \multicolumn{1}{c}{-} & \textit{1} & \textit{2}  & \textit{3} & \textit{4} & \multicolumn{1}{c|}{\textit{Avg.} $\downarrow$} & \textit{1} & \textit{2}  & \textit{3} & \textit{4} & \textit{Avg.} $\downarrow$ \\ \hline
JiGen~\citep{Carlucci_2019_CVPR} & \textit{ASD} & 8.55 & 14.09 & 11.35 & 12.57 & \multicolumn{1}{c|}{11.64} & 19.56 & 13.99 & 11.90 & 8.82 & 13.60 \\
DST~\citep{zhang2019unseen} & \textit{ASD} & 13.24 & \textbf{14.00} & 8.52 & 10.05 & \multicolumn{1}{c|}{11.45} & 24.42 & 12.89 & 10.91 & \textbf{7.05} & 13.82 \\
DoFE~\citep{wang2020dofe} & \textit{ASD} & 7.68 & 16.59 & 11.19 & 7.53 & \multicolumn{1}{c|}{10.75} & 16.94 & 13.87 & 9.59 & 7.24 & 11.91 \\ \hline
AFNN~(ours) & \textit{ASD} & \textbf{6.89} & 14.55 & \textbf{8.01} & \textbf{5.64} & \multicolumn{1}{c|}{\textbf{8.77}} & \textbf{14.64} & \textbf{12.58} & \textbf{9.02} & 7.19 & \textbf{10.86} \\ \hline
\multicolumn{11}{l}{\small *All the report scores are average quantity evaluations, and the lower ASD score denotes a better model performance.} \\
\end{tabular}}
\end{center}
\end{table*}

Table~\ref{tab:ASD} is the ASD evaluation results on four unseen domains. We can observe that our AFNN achieves superior performance than DoFE on all four, except for a slight performance decline in domain 2 for the optic-disk segmentation and domain 4 for optic-cup segmentation. Regarding the ASD as an edge-based evaluation metric, our AFNN significantly improves the edging smoothness than the segmentation methods of JiGen, DST, and DoFE. Meanwhile, more detailed comparisons are also provided in the discussion section. Overall, our AFNN achieves superior generalization performance than the compared baselines with large margins in both HD and ASD evaluations. 

\section{Discussion}
\label{Section5:Discussion}
\subsection{Ablation Study}
In the discussion section, we conduct an ablation study on AFNN's three modules: domain adaptor, feature-fusion network, and self-supervised multi-task learning module, which are abbreviated to ``Apt'', ``FF'', and ``MT'', respectively. We choose the DeepLabV3+ as the segmentation backbone for the baseline. Then, we study the contributions of each module by adding a selected module to the baseline model. All DSC, HD, and ASD results are the average scores of four unseen domains. 

\begin{table*}[!h]
\centering
\caption{Ablation study of AFNN.}
\label{tab:AblationStudy}
\begin{tabular}{rcccccccccccc}
\hline
\textbf{Methods} & \textbf{Adp} & \textbf{FF} & \textbf{MT} & \multicolumn{3}{c}{\textbf{DSC $\uparrow$}} & \multicolumn{3}{c}{\textbf{HD $\downarrow$}} & \multicolumn{3}{c}{\textbf{ASD $\downarrow$}} \\ \hline
\textbf{Type} & - & - & \multicolumn{1}{c|}{-} & \textit{OC} & \textit{OD} & \multicolumn{1}{c|}{\textit{Avg.}} & \textit{OC} & \textit{OD} & \multicolumn{1}{l|}{\textit{Avg.}} & \textit{OC} & \textit{OD} & \multicolumn{1}{l}{\textit{Avg.}} \\ \hline
\textbf{Baseline} &  &  & \multicolumn{1}{c|}{} & 0.7860 & 0.9148 & \multicolumn{1}{c|}{0.8504} & 29.95 & 22.33 & \multicolumn{1}{c|}{26.14} & 16.82 & 12.71 & 14.77 \\
\textbf{} & \checkmark &  & \multicolumn{1}{c|}{} & 0.8018 & 0.9287 & \multicolumn{1}{c|}{0.8652} & 26.68 & 19.48 & \multicolumn{1}{c|}{23.08} & 14.33 & 10.56 & 12.45 \\
\textbf{} &  & \checkmark & \multicolumn{1}{c|}{} & 0.8116 & 0.9204 & \multicolumn{1}{c|}{0.8660} & 25.94 & 19.36 & \multicolumn{1}{c|}{22.65} & 13.96 & 11.10 & 12.53 \\
\textbf{} &  &  & \multicolumn{1}{c|}{\checkmark} & 0.8018 & 0.9304 & \multicolumn{1}{c|}{0.8661} & 26.98 & 18.20 & \multicolumn{1}{c|}{22.59} & 14.44 & 9.97 & 12.21 \\ \hline
\textbf{AFNN} & \checkmark & \checkmark & \multicolumn{1}{c|}{\checkmark} & \textbf{0.8562} & \textbf{0.9393} & \multicolumn{1}{c|}{\textbf{0.8977}} & \textbf{21.17} & \textbf{17.43} & \multicolumn{1}{c|}{\textbf{19.3}} & \textbf{10.86} & \textbf{8.77} & \textbf{9.82} \\ \hline
\multicolumn{13}{l}{\small * The baseline is the DeepLabV3+ without data augmentation. ``$\checkmark$'' denotes the baseline with this module.} \\
\multicolumn{13}{l}{\small ** ``Adp'' is the domain adaptor, ``FF'' is the feature-fusion network, and ``MT'' is the self-supervised multi-task learning.} \\
\end{tabular}
\end{table*}

From the ablation study results of Table~\ref{tab:AblationStudy}, we observe that the baseline with any module can achieve a clear improvement in all evaluation metrics. In detail, the baseline with ``FF'' module gets fewer improvements than the baseline with ``Adp'' or ``MT'' module in DSC and ASD evaluation criteria, which indicates the stable inputs and self-supervised multi-task learning provides more help to model performance than the feature-fusion network in domain generalization task. The baseline with ``MT'' module achieves better performance than the other two settings, indicating that the self-supervised multi-task learning module is essential for AFNN in the low-data tasks. Overall, AFNN simultaneously with three modules achieves significant improvements over the settings with any single module in all evaluations.

\subsection{Domain-Adaptive Learning}
The domain adaptor module helps AFNN perform adaptive learning across different source domains, which maps all the input data into a common distribution to provide stable inputs for its neural networks. To verify the effectiveness of adaptive learning in AFNN, we conduct comparisons from two aspects: the data distribution and the feature visualization. The data distribution comparison explores the data's overall changes before and after the domain adaptor. Furthermore, the comparison of feature visualization explores the individual changes under adaptive learning.

\begin{figure}[!ht]
\centering
\includegraphics[width=\linewidth]{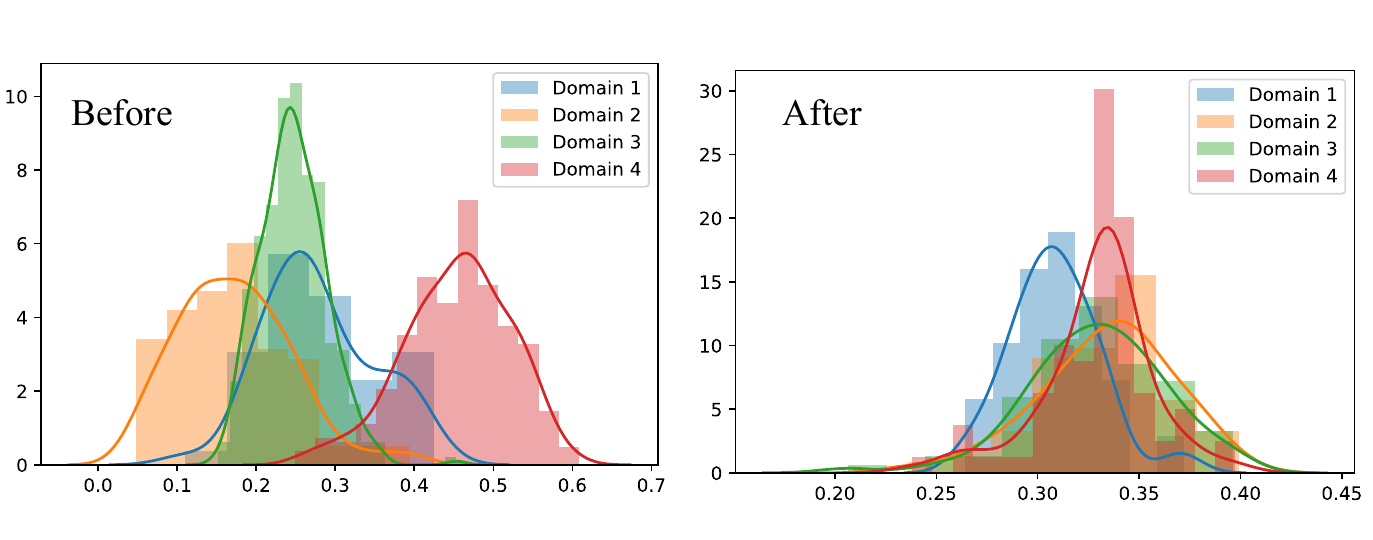}
\caption{The distribution comparison of domain-adaptive learning. The left sub-figure is the distribution of the raw data before adaptive learning, and the right sub-figure is the distribution of the raw data after adaptive learning.}
\label{fig8_AdaptiveDistribution}
\end{figure}

Figure~\ref{fig8_AdaptiveDistribution} shows the data distribution comparisons of raw data (before) and adapted data (after) in domain-adaptive learning. From the comparisons, we can observe that different domains have distinct distributions. All the domains follow the Gaussian distribution with an exception for domain 3. In addition, domain 4 has a large gap with the other domains in data distributions. With the help of domain adaptor, all the domain gaps are significantly reduced after domain-adaptive learning with the normalized distributions (Figure~\ref{fig8_AdaptiveDistribution} right) than the distributions of raw data (Figure~\ref{fig8_AdaptiveDistribution} left). While only domain 3 has a slight distribution shift from the rest domains, because its original distribution is not the Gaussian-like distribution in the normalized adaptation. Overall, all the data distributions are transformed into a common distribution. 

Besides the comparisons on data distributions, we also employ the T-SNE to visualize the changes of individual samples in domain-adaptive learning. All domains randomly select the same number of samples in T-SNE visualization.

\begin{figure}[!ht]
\centering
\includegraphics[width=\linewidth]{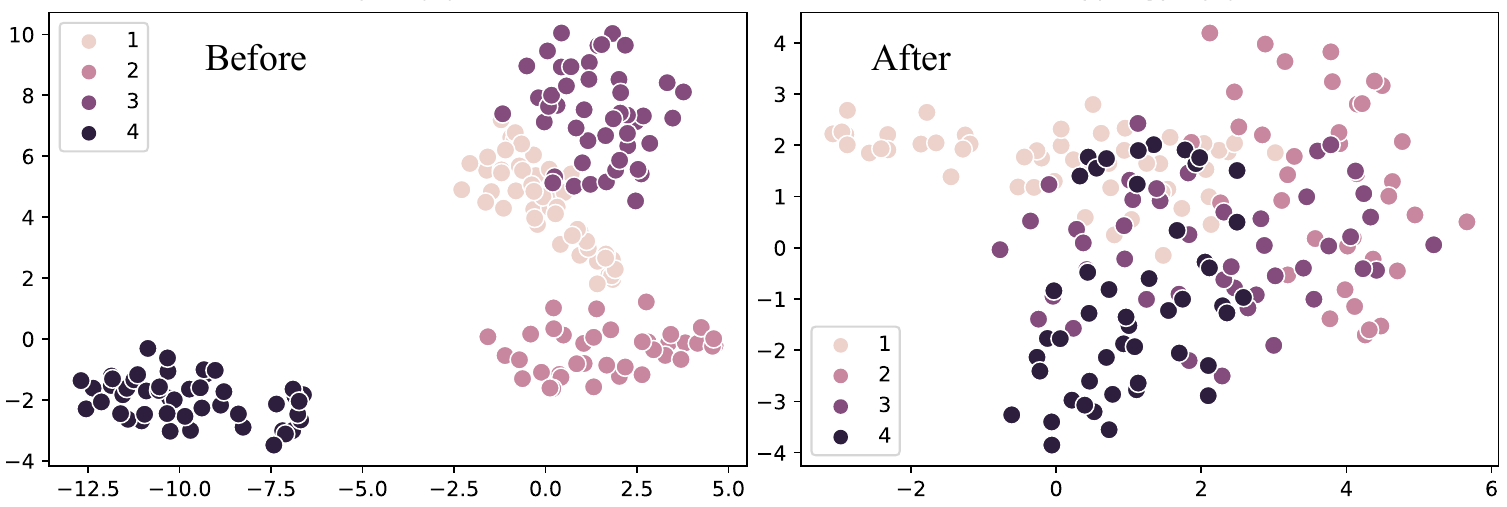}
\caption{Feature comparison of domain-adaptive learning with T-SNE visualization. The dots with different colors denote the samples coming from different domains. The left sub-figure presents the feature visualization of raw data, and the right sub-figure presents the feature visualization after domain-adaptive learning.}
\label{fig9_TSNE}
\end{figure}

Figure~\ref{fig9_TSNE} shows the feature comparison of raw data (left) and adapted data (right) in the domain-adaptive learning. From the left sub-figure, we can observe that a linear classification boundary can clearly separate data samples from different domains. Moreover, the distances of the samples among domains 1, 2, and 3 are shorter than their distance to domain 4, which reveals that domain 4 has a domain gap with other domains in the data distributions (Figure~\ref{fig8_AdaptiveDistribution}). After the domain-adaptive learning, all the data samples are mapped into a non-linear classifiable feature space where all the features from different domains follow a same standard distribution. In other words, all the samples from different domains are mapped with relatively equidistant distances in the new feature space, which provides stable inputs for the deep network. Overall, the comparisons of total distributions and individual samples verify that our domain adaptor helps AFNN conduct adaptive learning effectively across different domains.

\subsection{Multi-task Learning}
Multi-task learning is a popular solution for improving the learning capability of deep models in training sources with small datasets. It learns external knowledge from the training sources and enhances the model's representation capability on domain generalization tasks. In this section, we explore two learning paradigms of multi-task learning: parallel multi-task learning and sequential multi-task learning. 
Although the two multi-task learning paradigms can be trained for the same target task, the different paradigm leads to different learning trajectory. Specifically, sequential multi-task learning is similar to the paradigm of pre-training by training associate-task and target-task sequentially. In contrast, parallel multi-task training is a paradigm of training associate-task and target-task simultaneously. We compare the model performance on the target domain to explore the difference between two multi-task training paradigms. 

\begin{figure}[!ht]
\centering
\includegraphics[width=\linewidth]{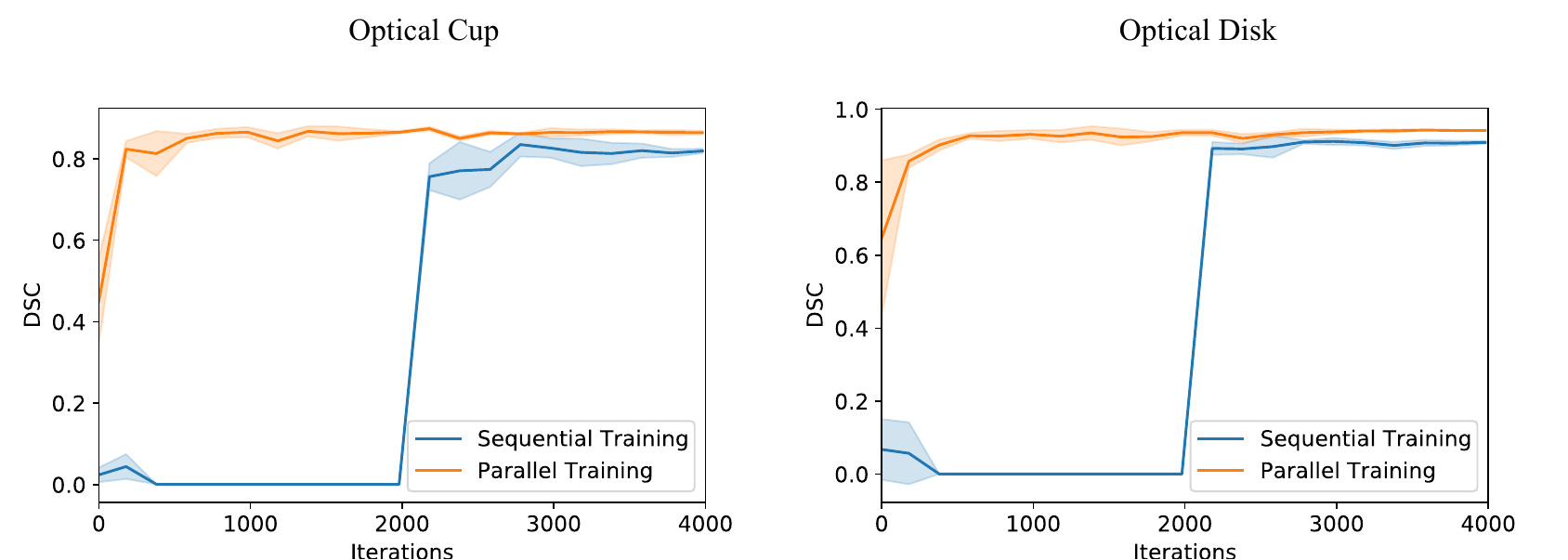}
\caption{Quantity comparisons of parallel multi-task learning and sequential multi-task learning. The left sub-figure is the DSC score comparison on optical-cup segmentation task, and the right sub-figure is the DSC score comparison on optical-disk segmentation task.}
\label{fig7_SeqML}
\end{figure}

\begin{figure*}[!t]
\centering
\includegraphics[width=\linewidth]{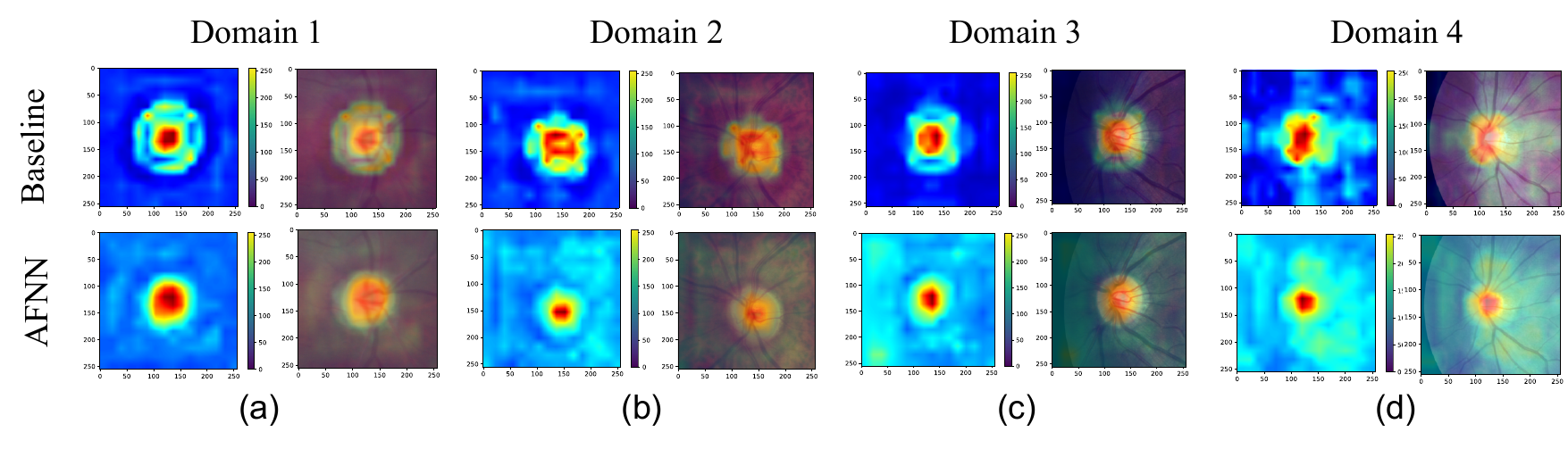}
\caption{Attention map visualization on different unseen domains. Each sub-figure illustrates the attention map (left) and overlapped fundus image (right). The first row is the visualization results of the baseline, the second row is the visualization results of AFNN.}
\label{fig13_AttentionComparison}
\end{figure*}

From the comparison results of Figure~\ref{fig7_SeqML},  we observe that the parallel multi-task learning paradigm achieves a superior DSC performance than the sequential multi-task learning paradigm on both optic-cup and optic-disk segmentation tasks. Surprisingly, the performance curves are nearly zero for the target task in the sequential multi-task learning paradigm through its whole associate-task training process, which indicates the target segmentation task only benefits from good initialization at the starting epochs and benefits little from the associate task training. In other words, sequential multi-task learning leads the learning trajectory away from the target learning trajectory. In contrast, the performance curves increase stably in the parallel multi-task learning paradigm, which indicates the trajectory of the associate task always keeping along with the learning trajectory of the target task.

\subsection{Feature-fusion in AFNN}
In this section, we visualized the feature attention maps with the convolution attention map approach~\citep{komodakis2017paying}. Similarly, we also choose the DeepLabV3+ as the baseline, whose segmentation framework is only with vertical feature-fusion. In contrast, our feature-fusion neural network improves AFNN's representation capability with vertical and horizontal feature-fusion for the domain generalization task. All the fundus image samples are randomly selected from different domains, and comparison results are summarized in Figure~\ref{fig13_AttentionComparison}.

From the comparisons of the attention map, we can observe that both the baseline and AFNN give attention to the target segmentation areas correctly as the attention maps cover the target areas of both optic-cup and optic-disk. However, our AFNN gains more precise and smooth edges on the attention maps than the baseline method with the help of the feature-fusion network. More importantly, AFNN's attention maps achieve the higher attention scores on the optic-cup area than the baseline results, which is consistent with the AFNN's motivation to focus on the challenging optic-cup segmentation task with enlarged loss weight. Moreover, most of our AFNN attention maps provide the predictions with smooth circle-like edges in the optic-disk segmentation task. In contrast, the attention maps of the baseline methods are predicted with blurred edges in their predicted results. 

\subsection{Segmentation Visualization}
\begin{figure*}[!t]
\centering
\includegraphics[width=0.8\linewidth]{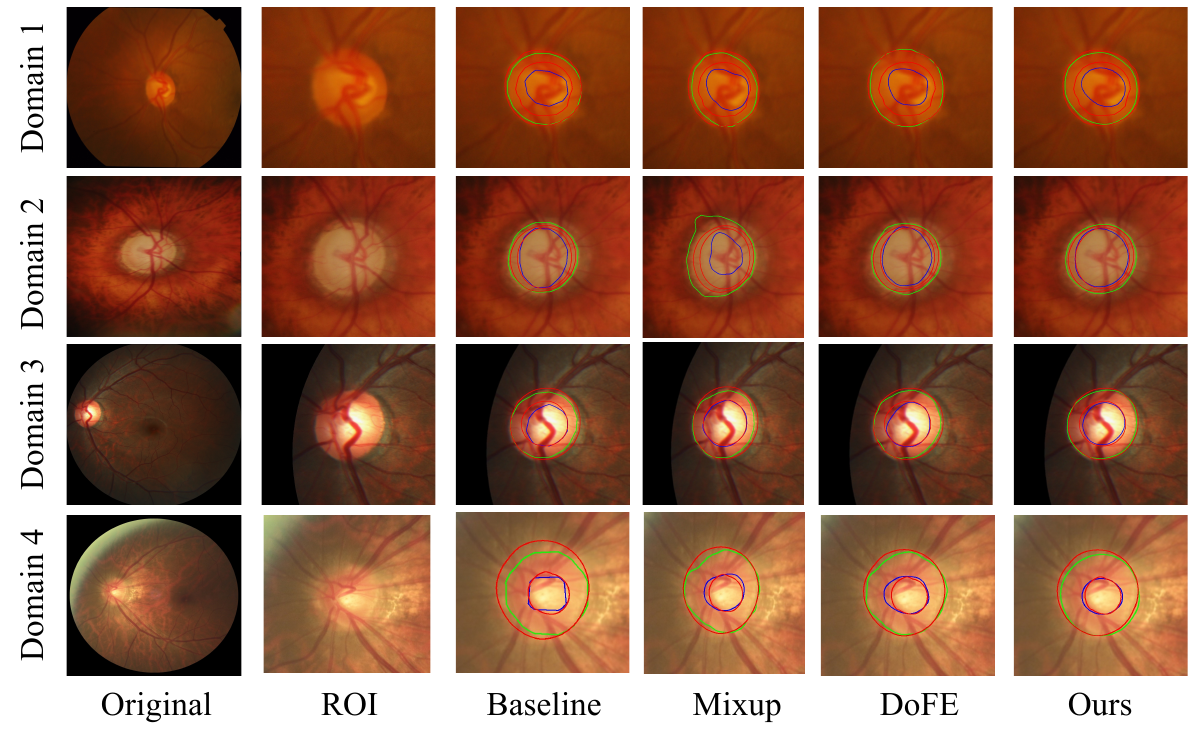}
\caption{The glaucoma segmentation result comparisons on unseen domains. All the original fundus images are randomly selected and then resized to the same resolution. The sub-figures in the same row denote the samples from the same domain, and the sub-figures in the same column denote the prediction results of the same segmentation method. The green and blue edges are the ground truth of optic disk and optic cup, correspondingly. The red edges are the predictions of different glaucoma segmentation methods.}
\label{fig10_restCmp}
\end{figure*}

As can be seen from the first row of Figure~\ref{fig10_restCmp}, the fundus images from different domains exhibit significant differences in lightness, resolutions, contrast, and view angles. To evaluate the generalization performance of different segmentation methods, we select the DeepLabV3+ as the baseline, Mixup and DoFE as the preventative segmentation methods. All the comparison results are summarized in Figure~\ref{fig10_restCmp}.

Specifically, the fundus image of domain 1 has blurred optic-cup edges, which leads to model performance degradation in the optic-cup segmentation task. From the result comparison of domain 2, Mixup wrongly predicts the optic-disk edges to the optic-disk area, which is caused by Mixup's interpolating manner. All the segmentation methods perform well on domain 3. Nevertheless, DoFE performs sensitively to the complex lightness and contrast changes on the last domain 4, which wrongly predicts the blood vessel to the optic-cup area. The performance of baseline is dramatically declined on the optic-disk segmentation results. This is because the optic disk edge is even more blurry on domain 4 than on domain 1. Our AFNN always presents the best performances on all unseen domains with the help of adaptive learning, feature-fusion, and multi-task learning modules. More importantly, our AFNN also delivers the glaucoma segmentation results with smooth edges, which leads to AFNN achieving superior performance on all edge-based HD and ASD metrics. 

\subsection{Different Training Sources}
To investigate the impacts of different training sources in the low-data glaucoma segmentation task, we evaluate the model performance that trained on different training sources. We have three experimental settings (one, two, and three source datasets) with respect to one out of four public glaucoma datasets for evaluation. All the reported results are the averaged performance of different training combination settings. Here, we take the two training sources setting for example. Firstly, one target domain (domain 1) is selected as the evaluation domain out of four public datasets. Then, two training sources (e.g., domains 2 and 3) are randomly selected as the training dataset from the rest three available sources (domains 2, 3, and 4). Consequently, we have three available training combinations in this setting(domain 2$\&$3, domain 2$\&$4, domain 3$\&$4). At last, the reported results are the average of three available training combinations, which are all evaluated on the unseen domain (domain 1).

\begin{figure}[!ht]
\centering
\includegraphics[width=\linewidth]{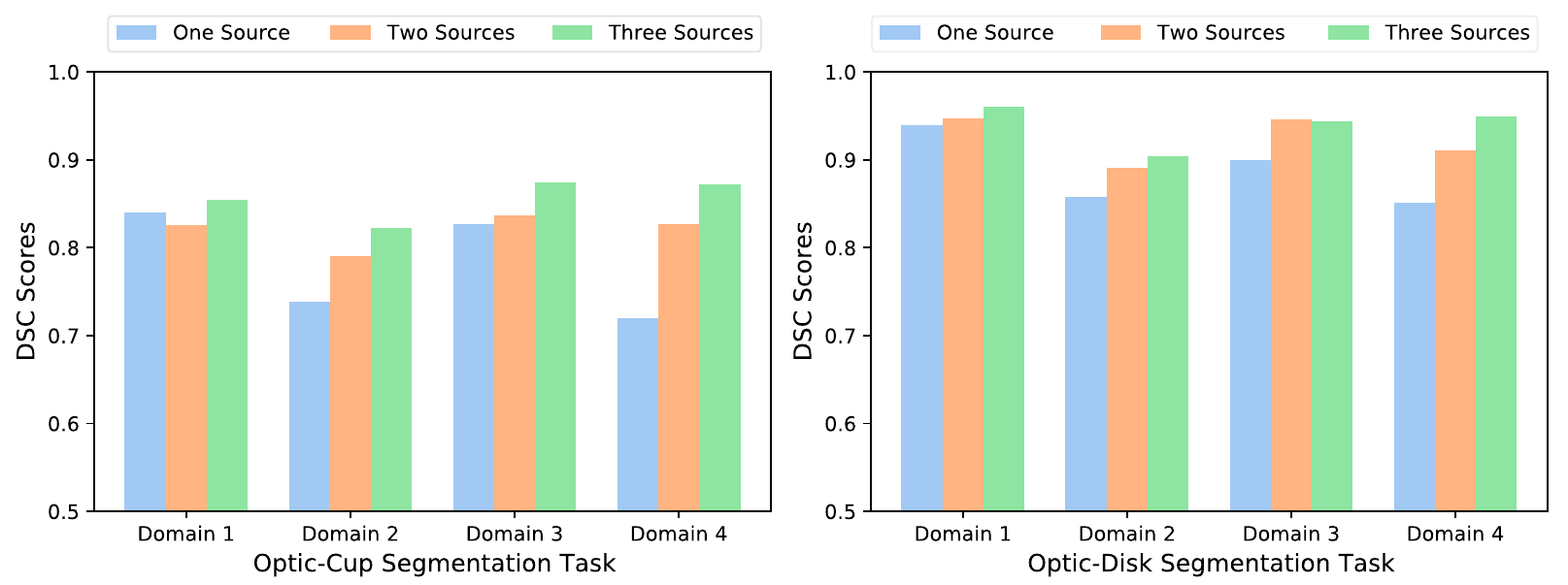}
\caption{DSC evaluation on different training source sittings. The left and right sub-figures are the optic-cup and optic-disk segmentation tasks separately. The X-axis is the target evaluation domains, and the Y-axis is the average DSC performance on different training settings.}
\label{fig11_sourceComparison}
\end{figure}

The different settings' DSC results are reported in Figure~\ref{fig11_sourceComparison}. We can only find that the settings with more training sources always achieve a better performance than the settings with fewer training sources on both optic-cup and optic-disk tasks. Except for the domain 1 evaluation in the optic-cup segmentation task, the model performance trained on two sources declines more than that trained on one source. Moreover, all the optic-disk segmentation settings achieve a better DSC performance than the optic-cup segmentation settings, indicating the optic-cup segmentation tasks are more challenging than the optic-disk segmentation tasks. 

\begin{figure}[!ht]
\centering
\includegraphics[width=\linewidth]{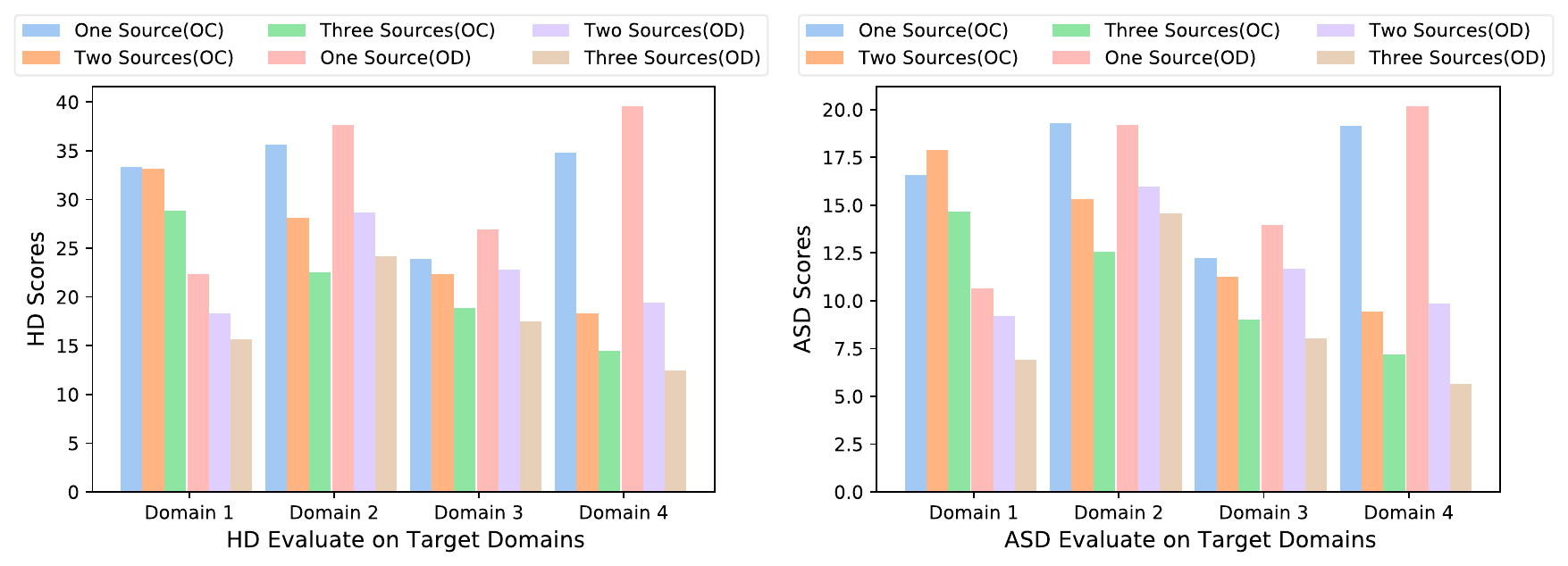}
\caption{HD and ASD evaluation on different training sources. The left and right sub-figures are the HD and ASD evaluation results separately. ``OC'' and ``OD'' correspondingly denote optic-cup segmentation tasks and optic-disk segmentation tasks. The X-axis is the target domain, and the Y-axis is the mean of HD or ASD performance in different training settings.}
\label{fig12_sourceHDComparison}
\end{figure}

Figure~\ref{fig12_sourceHDComparison} shows the HD and ASD evaluation results in different training settings. The comparison shows that the settings with more training sources generally perform better in the optic-cup and optic-disk segmentation tasks with lower HD and ASD scores. The only exception is domain 1 ASD evaluation in the optic-cup segmentation task, where the two source training setting gets a higher ADS score than the one source training setting. Compared with DSC performance, this abnormal ASD score is caused by the performance decline of DSC evaluation in domain 1. In contrast, the optic-disk DSC performance decline in domain 1 makes no impact on both HD and ASD evaluations that the setting with more training sources always gets the lower scores. This indicates that our AFNN can consistently predict the smooth edges. Even its area-related DSC performance gets declines in its optic-cup segmentation task. 

\section{Conclusion}
In this work, we present a novel method named adaptive feature-fusion neural network (AFNN) for glaucoma segmentation on unseen domains. Our AFNN contains three modules: domain adaptor, feature-fusion network, and self-supervised multi-task learning module. Specifically, the domain adaptor provides normalized inputs for stable training across different sources. The feature-fusion network and the self-supervised multi-task learning improve model representation capability for domain generalization tasks. Moreover, the staged optimize strategy and weighted dice loss further improve model performance on low-data glaucoma segmentation tasks. Extensive experiments demonstrate that our AFNN is superior to the existing glaucoma segmentation methods on four public datasets. In future work, we will evaluate our AFNN more on other data-limited medical image analysis tasks.

\bibliographystyle{unsrtnat}
\bibliography{mybib}

\section*{Supplementary Material}

\textbf{Quantity Test for Glaucoma Segmentation.} We provide the detailed results for quantity evaluations on metrics of DSC (Table~\ref{tab:appendix1_DSC}), HD (Table~\ref{tab:appendix2_HD}) and ASD (Table~\ref{tab:appendix3_ASD}). We conduct three experiments on different domains in the quantity evaluation, and the ``T'' denotes the index of quantity evaluation.

\setcounter{table}{0}
\begin{table}[!ht]
\caption{DSC Quantity Evaluation.}
\centering
\scalebox{0.75}{
\label{tab:appendix1_DSC}
\begin{tabular}{rcccccccc}
\hline
\textbf{Task} & \multicolumn{4}{c|}{\textbf{DSC} (Optic-Cup)} & \multicolumn{4}{c}{\textbf{DSC} (Optic-Disk)} \\ \hline
Domains&\textit{1}&\textit{2}& \textit{3} & \multicolumn{1}{c|}{\textit{4}} & \textit{1} & \textit{2} & \textit{3} & \textit{4} \\ \hline
\textit{T1}  & 0.8529  & 0.8097  & 0.8727  & 0.8727 & 0.9573  & 0.9029  & 0.9482 & 0.9507 \\
\textit{T2}  & 0.8549  & 0.8384  & 0.8693  & 0.8727 & 0.9621  & 0.9024  & 0.9417 & 0.9504 \\
\textit{T3}  & 0.8569  & 0.8206  & 0.8806  & 0.8725 & 0.9614  & 0.9071  & 0.9413 & 0.9467 \\ \hline
\textbf{Avg.} & 0.8549  & 0.8229  & 0.8742  & 0.8726 & 0.9602  & 0.9041  & 0.9437 & 0.9492 \\ \hline
\end{tabular}}
\end{table}

\begin{table}[!ht]
\caption{HD Quantity Evaluation}
\centering
\scalebox{0.85}{
\label{tab:appendix2_HD}
\begin{tabular}{rcccccccc}
\hline
\textbf{Task} & \multicolumn{4}{c|}{\textbf{HD} (Optic-Cup)}                           & \multicolumn{4}{c}{\textbf{HD} (Optic-Disk)}      \\ \hline
Domains       & \textit{1} & \textit{2} & \textit{3} & \multicolumn{1}{c|}{\textit{4}} & \textit{1} & \textit{2} & \textit{3} & \textit{4} \\ \hline
\textit{T1}   & 28.55 & 24.26 & 19.09 & 14.17 & 16.86 & 24.82 & 16.20 & 11.61 \\
\textit{T2}   & 29.02 & 20.96 & 19.34 & 14.20 & 14.99 & 23.93 & 18.04 & 12.41 \\
\textit{T2}   & 29.06 & 22.29 & 18.25 & 14.93 & 15.01 & 23.72 & 18.13 & 13.46 \\ \hline
\textbf{Avg.} & 28.87 & 22.50 & 18.89 & 14.43 & 15.62 & 24.16 & 17.46 & 12.49 \\ \hline
\end{tabular}}
\end{table}

\begin{table}[!ht]
\caption{ASD Quantity Evaluation}
\begin{center}
\scalebox{0.87}{
\label{tab:appendix3_ASD}
\begin{tabular}{rcccccccc}
\hline
\textbf{Task} & \multicolumn{4}{c|}{\textbf{ASD} (Optic-Cup)}                          & \multicolumn{4}{c}{\textbf{ASD} (Optic-Disk)}     \\ \hline
Domains       & \textit{1} & \textit{2} & \textit{3} & \multicolumn{1}{c|}{\textit{4}} & \textit{1} & \textit{2} & \textit{3} & \textit{4} \\ \hline
\textit{T1}   & 14.96 & 13.26 & 9.07 & 7.17 & 7.36 & 14.30 & 7.49 & 5.46 \\
\textit{T2}   & 14.58 & 11.21 & 9.35 & 7.16 & 6.60 & 13.21 & 8.25 & 5.51 \\
\textit{T2}   & 14.39 & 13.27 & 8.65 & 7.24 & 6.71 & 16.15 & 8.29 & 5.95 \\ \hline
\textbf{Avg.} & 14.64 & 12.58 & 9.02 & 7.19 & 6.89 & 14.55 & 8.01 & 5.64 \\ \hline
\end{tabular}}
\end{center}
\end{table}

\begin{table*}[!t]
\begin{center}
\caption{Ablation study: DSC performance.}
\label{tab:appedix_ablationDSC}
\resizebox{\textwidth}{!}{ %
\begin{tabular}{cccccccccccc|c}
\hline
\textbf{Methods} & \textbf{Adp} & \textbf{FF} & \textbf{MT} & \multicolumn{4}{c}{\textbf{DSC} (Optic-Cup)} & \multicolumn{4}{c}{\textbf{DSC} (Optic-Disk)} &  \\ \hline
Domains & - & - & \multicolumn{1}{c|}{-} & \textit{1} & \textit{2} & \textit{3} & \multicolumn{1}{c|}{\textit{4}} & \textit{1} & \textit{2} & \textit{3} & \textit{4} & \textit{Avg.} \\ \hline
Baseline &  &  & \multicolumn{1}{c|}{} & 0.7661 & 0.7884 & 0.7508 & \multicolumn{1}{c|}{0.8386} & 0.9272 & 0.9097 & 0.8871 & 0.9349 & 0.8454 \\
 & \checkmark &  & \multicolumn{1}{c|}{} & 0.7927 & 0.7973 & 0.7718 & \multicolumn{1}{c|}{0.8453} & 0.9430 & 0.9098 & 0.9160 & 0.9459 & 0.8652 \\
 &  & \checkmark  & \multicolumn{1}{c|}{} & 0.7763 & 0.8198 & 0.7957 & \multicolumn{1}{c|}{0.8546} & 0.9442 & 0.9186 & 0.8862 & 0.9327 & 0.8660 \\
 &  &  & \multicolumn{1}{c|}{\checkmark } & 0.7784 & 0.7955 & 0.7807 & \multicolumn{1}{c|}{0.8525} & 0.9355 & 0.9117 & 0.9318 & 0.9427 & 0.8661 \\ \hline
AFNN & \checkmark  & \checkmark  & \multicolumn{1}{c|}{\checkmark} & 0.8549 & 0.8229 & 0.8742 & \multicolumn{1}{c|}{0.8726} & 0.9602 & 0.9041 & 0.9437 & 0.9492 & 0.8977 \\ \hline
\end{tabular}}
\end{center}
\end{table*}

\begin{table*}[!h]
\caption{Ablation study: HD performance}
\centering
\label{tab:appedix_ablationHD}
\begin{tabular}{cccccccccccc|c}
\hline
\textbf{Methods} & \textbf{Adp} & \textbf{FF} & \textbf{MT} & \multicolumn{4}{c}{\textbf{HD} (Optic-Cup)} & \multicolumn{4}{c}{\textbf{HD} (Optic-Disk)} & \\ \hline
Domains & - & - & \multicolumn{1}{c|}{-} & \textit{1} & \textit{2} & \textit{3} & \multicolumn{1}{c|}{\textit{4}} & \textit{1} & \textit{2} & \textit{3} & \textit{4} &  \textit{Avg.}\\ \hline
Baseline &  &  & \multicolumn{1}{c|}{} & 38.91 & 31.16 & 31.60 & \multicolumn{1}{c|}{18.13} & 20.19 & 27.89 & 25.82 & 15.43 & 26.14 \\
 & \checkmark  &  & \multicolumn{1}{c|}{} & 37.48 & 23.78 & 27.93 & \multicolumn{1}{c|}{17.54} & 17.79 & 23.45 & 23.33 & 13.36 & 23.08 \\
 &  & \checkmark  & \multicolumn{1}{c|}{} & 39.89 & 22.25 & 25.02 & \multicolumn{1}{c|}{16.61} & 16.75 & 20.08 & 26.42 & 14.17 & 22.65 \\
 &  &  & \multicolumn{1}{c|}{\checkmark } & 38.96 & 23.90 & 27.50 & \multicolumn{1}{c|}{17.58} & 18.42 & 22.25 & 19.01 & 13.11 & 22.59 \\ \hline
AFNN &\checkmark   & \checkmark  & \multicolumn{1}{c|}{\checkmark } & 28.87 & 22.50 & 18.89 & \multicolumn{1}{c|}{14.43} & 15.62 & 24.16 & 17.46 & 12.49 & 19.3 \\ \hline
\end{tabular}
\end{table*}

\begin{table*}[!h]
\caption{Ablation study: ASD performance}
\label{tab:appedix_ablationASD}
\begin{tabular}{cccccccccccc|c}
\hline
\textbf{Methods} & \textbf{Adp} & \textbf{FF} & \textbf{MT} & \multicolumn{4}{c}{\textbf{ASD} (Optic-Cup)} & \multicolumn{4}{c}{\textbf{ASD} (Optic-Disk)} & \\ \hline 
Domains & - & - & \multicolumn{1}{c|}{-} & \textit{1} & \textit{2} & \textit{3} & \multicolumn{1}{c|}{\textit{4}} & \textit{1} & \textit{2} & \textit{3} & \textit{4} & \textit{Avg.} \\ \hline
Baseline &  &  & \multicolumn{1}{c|}{} & 23.48 & 17.66 & 17.27 & \multicolumn{1}{c|}{8.87} & 12.32 & 14.63 & 15.61 & 8.29 & 14.77 \\
 & \checkmark  &  & \multicolumn{1}{c|}{} & 20.98 & 12.90 & 14.99 & \multicolumn{1}{c|}{8.44} & 9.70 & 14.77 & 11.71 & 6.04 & 12.45 \\
 &  & \checkmark  & \multicolumn{1}{c|}{} & 22.48 & 11.72 & 13.77 & \multicolumn{1}{c|}{7.88} & 9.50 & 11.79 & 15.79 & 7.33 & 12.53 \\
 &  &  & \multicolumn{1}{c|}{\checkmark } & 22.40 & 12.61 & 14.65 & \multicolumn{1}{c|}{8.09} & 10.96 & 12.93 & 9.62 & 6.35 & 12.21 \\ \hline
AFNN & \checkmark  & \checkmark  & \multicolumn{1}{c|}{\checkmark} & 14.64 & 12.58 & 9.02 & \multicolumn{1}{c|}{7.19} & 6.89 & 14.55 & 8.01 & 5.64 & 9.82 \\ \hline
\end{tabular}
\end{table*}

\textbf{Ablation Study Details.} We present the domain specific results in ablation study. The baseline is DeepLabV3+, which also is the baseline on glaucoma segmentation task. We study the performance improvements on three main modules of AFNN: domain adaptor (``Apt''), feature-fusion neural network (``FF'') and self-supervised multi-task learning (``MT''). Due to the space limitation, different evaluation metric are reported in separated tables (DSC in Table~\ref{tab:appedix_ablationDSC}, HD in Table~\ref{tab:appedix_ablationHD}, and ASD in Tables~\ref{tab:appedix_ablationASD}).

\end{document}